\pdfoutput=1
\documentclass[letterpaper]{article} 
\usepackage{aaai2026}  
\usepackage{times}  
\usepackage{helvet}  
\usepackage{courier}  
\usepackage[hyphens]{url}  
\usepackage{graphicx} 
\usepackage{natbib}  
\usepackage{caption} 
\usepackage{algorithm}
\usepackage{algorithmic}
\usepackage{newfloat}
\usepackage{listings}
\usepackage{amsmath, amssymb}
\usepackage{multirow}
\usepackage[table]{xcolor}
\DeclareCaptionStyle{ruled}{labelfont=normalfont,labelsep=colon,strut=off} 
\floatstyle{ruled}
\newfloat{listing}{tb}{lst}{}
\floatname{listing}{Listing}
\title{StegaVAR: Privacy-Preserving Video Action Recognition via Steganographic Domain Analysis}
\author{
    Lixin Chen\textsuperscript{\rm 1}\equalcontrib,
    Chaomeng Chen\textsuperscript{\rm 1,\rm 2,\rm 6}\equalcontrib,
    Jiale Zhou\textsuperscript{\rm 4,\rm 5},
    Zhijian Wu\textsuperscript{\rm 5},
    Xun Lin\textsuperscript{\rm 3}\thanks{Corresponding author.}
}
\affiliations{
    \textsuperscript{\rm 1}School of Computing and Information Technology, Great Bay University\\
    \textsuperscript{\rm 2}Tsinghua Shenzhen International Graduate School, Tsinghua University\\
    \textsuperscript{\rm 3}School of Computer Science and Engineering, Beihang University\\
    \textsuperscript{\rm 4}College of Science and Technology, Zhejiang University\\
    \textsuperscript{\rm 5}School of Engineering, Westlake University\\
    \textsuperscript{\rm 6}Dongguan Key Laboratory for Intelligence and Information Technology\\

    clxhsa@outlook.com, linxun@buaa.edu.cn
}

\begin{document}

\maketitle

\begin{abstract}
Despite the rapid progress of deep learning in video action recognition (VAR) in recent years, privacy leakage in videos remains a critical concern. Current state-of-the-art privacy-preserving methods often rely on anonymization. These methods suffer from (1) low concealment, where producing visually distorted videos that attract attackers’ attention during transmission, and (2) spatiotemporal disruption, where degrading essential spatiotemporal features for accurate VAR. To address these issues, we propose StegaVAR, a novel framework that embeds action videos into ordinary cover videos and directly performs VAR in the steganographic domain for the first time. Throughout both data transmission and action analysis, the spatiotemporal information of hidden secret video remains complete, while the natural appearance of cover videos ensures the concealment of transmission. Considering the difficulty of steganographic domain analysis, we propose \textit{S}ecret Spatio-\textit{Te}mporal \textit{P}romotion (\textit{STeP}) and \textit{Cro}ss-Band \textit{D}ifference \textit{A}ttention (\textit{CroDA}) for analysis within the steganographic domain. STeP uses the secret video to guide spatiotemporal feature extraction in the steganographic domain during training. CroDA suppresses cover interference by capturing cross-band semantic differences. Experiments demonstrate that StegaVAR achieves superior VAR and privacy-preserving performance on widely used datasets. Moreover, our framework is effective for multiple steganographic models. The codes will be released soon.
\end{abstract}


\vspace{-0.95em}
\section{Introduction}
\vspace{-0.15em}

Video action recognition (VAR) aims to automatically identify the body's movement patterns and behaviors from videos \cite{sunTPAMI2023}. This technology is increasingly applied in real-world scenarios like video surveillance, which rely on extensive data transmission and cloud-based analysis \cite{PittalugaTPAMI2017, fitwiICCCN2020, Deng_2023_ICCV, xie2024fusionmamba}. However, this process introduces a significant privacy concern \cite{SPAct, chen2024enforcing, lin2024safeguarding, chen2025pfdp}, as sensitive attributes such as gender, race, appearance, and the actions themselves can be exposed when videos are uploaded to remote servers. Consequently, privacy-preserving VAR solutions are urgently needed.

Initial attempts at privacy preservation, such as extreme downsampling \cite{daiICIP2015, liuICAICE2020, ryooAAAI2017, srivasstavMICCAI2019} or naive region obfuscation \cite{Ren_2018_ECCV, zhangELECTRONICS2021}, proved insufficient, as they severely degrade VAR accuracy. With the development of adversarial learning, anonymization-based methods have become the dominant paradigm. These methods, employing adversarial \cite{wuTPAMI2022, STPrivacy} and self-supervised \cite{SPAct, Aslam_2025_CVPR} frameworks, aim to remove sensitive information while retaining utility. Despite their advancements, existing anonymization techniques are designed to protect privacy, inadvertently introduce new risks, and performance ceilings. \textit{How precisely does this dilemma arise?}

\begin{figure*}[t]   
	\centering
	\includegraphics[width=\linewidth,scale=0.85]{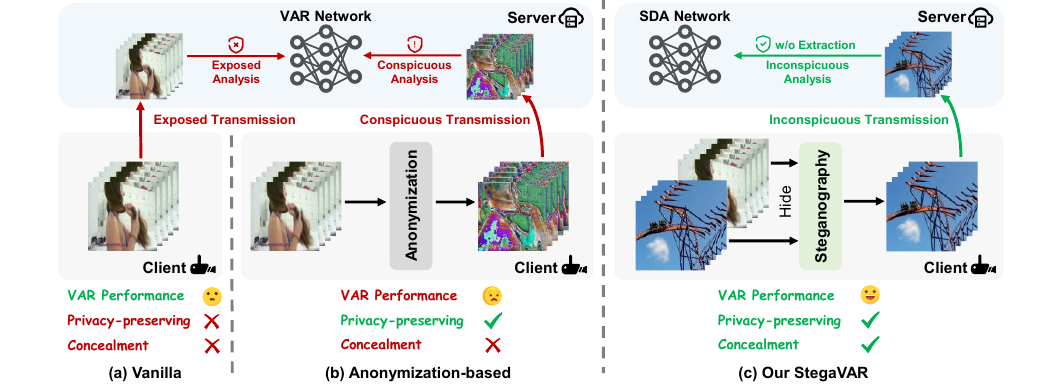}
    \vspace{-1.85em}
	\caption{Different client-server VAR framework: (a) vanilla, (b) anonymization-based, (c) our proposed StegaVAR framework. StegaVAR can achieve accurate VAR without attracting the attention of attackers.}
	\label{intro}
\vspace{-0.85em}
\end{figure*}

We argue this failure stems from two critical, unaddressed problems inherent to the anonymization process: \textbf{(1) Low Concealment.} Anonymization techniques can easily produce visually distorted videos. These alterations create visual artifacts that distinguish them from the torrent of natural video data. This conspicuousness is a security flaw rather than a benign side effect. It acts as a red flag to network adversaries, signaling that the content is sensitive enough to warrant protection. This signal leads to a \textit{cat-and-mouse game} of anonymization and de-anonymization, which may cause more aggressive server/client attack \cite{Rosberg_2023_ICCV, Andrea_sciadv_2024}. \textbf{(2) Spatiotemporal Disruption.} The process of anonymization is inherently destructive, which irrevocably corrupts the video's pixel data to obscure private information. This corruption often disrupts the fine-grained spatiotemporal relationships and high-frequency details, which are necessary for accurate VAR \cite{wangTPAMI2019, Feichtenhofer_2019_ICCV, tpami2}.

Drawing inspiration from recent advancements in video steganography \cite{guanTPAMI2022, Lu_2021_CVPR, HiNet, Li_2024_CVPR, LF_VSN, Deng_2023_ICCV}, we propose a novel framework for privacy-preserving video action recognition via steganographic domain analysis (StegaVAR), reframing the problem from one of ``editing'' a video to one of ``hiding'' it. StegaVAR protects privacy information in videos without attracting attackers' attention. As shown in Fig.~\ref{intro}, our approach conceals the secret video within a natural cover video to generate the stego video for server upload, then directly performs VAR in the steganographic domain without video extraction or decryption. Since secret information is primarily embedded in high-frequency components of the stego video \cite{HiNet}, these features remain subtle, while cover information introduces substantial interference that impedes discriminative feature extraction for action recognition. The subtlety of these features and interference from the cover video make it challenging to apply existing action models to the steganographic domain directly. We propose a Steganographic Domain Analysis network (SDANet) to solve the problem.

Within SDANet, we design two modules: \textbf{S}ecret Spatio-\textbf{Te}mporal \textbf{P}romotion (\textbf{STeP}) and \textbf{Cro}ss-Band \textbf{D}ifference \textbf{A}ttention (\textbf{CroDA}). Considering video steganography embeds secret videos into high-frequency components of cover videos, we decompose videos using Discrete Wavelet Transform \cite{hePR2012} and analyze different frequency bands separately. During training, STeP utilizes the secret video's high-frequency components to guide feature extraction in the stego video's spatial \cite{zhangTIP2023} and temporal dimensions. CroDA computes differences between frequency bands to suppress interference from the cover video \cite{tpami1, tpami3}, further enhancing performance with a shared position embedding to maintain temporal consistency across sub-bands. Overall, our proposed StegaVAR conceals privacy attributes in action videos without raising suspicion and minimally disrupts spatiotemporal features in secret videos, thereby achieving accurate privacy-preserving VAR. Our contributions are summarized as follows:

\begin{itemize}
    \item We propose the StegaVAR framework, a new paradigm for privacy-preserving VAR, which integrates video steganography with VAR for the first time, realizing accurate VAR without attracting attackers' attention.
    \item We design the STeP and CroDA modules to perform VAR directly in the steganographic domain, which fully leverages secret features in high-frequency components, protecting privacy while maintaining VAR performance.
    \item Our method achieves strong generalizability across multiple steganographic models in terms of VAR acc and concealment capability on six publicly available datasets.
\end{itemize}

\section{Related Work}

\subsubsection{Privacy-preserving Action Recognition.}

Early privacy-preserving action recognition methods relied on direct data manipulation, such as extreme downsampling to suppress sensitive features \cite{daiICIP2015, chou2018privacy, srivasstavMICCAI2019}. Obfuscation-based techniques use pretrained detectors to find and modify private regions, for instance by synthesizing fake faces \cite{Ren_2018_ECCV} or applying segmentation-guided blurring \cite{zhangELECTRONICS2021}. However, these methods are fundamentally limited. Downsampling severely degrades motion recognition performance, while obfuscation suffers from poor generalization to unseen privacy attributes, the impracticality of frame-level annotation, and the negative impact of modifications on downstream task performance. To better address the privacy-utility trade-off, research shifted towards end-to-end, learning-based methods. Initial works leveraged supervised adversarial learning to obfuscate features \cite{Wu_2018_ECCV, wuTPAMI2022}, which was later improved by the MaSS framework for selective attribute preservation \cite{chen2022mass}. STPrivacy designed a transformer-based model to remove action-irrelevant information at the video level \cite{STPrivacy}. More recently, self-supervised learning has eliminated the need for privacy labels, as demonstrated by methods that optimize privacy without annotations \cite{SPAct} or use contrastive learning to mitigate spatial privacy leakage \cite{Fioresi_2023_ICCV}. A penalty-based optimization algorithm was also introduced to further balance privacy and task performance \cite{Aslam_2025_CVPR}.

\begin{figure}[t]   
	\centering
	\includegraphics[width=\linewidth,scale=0.85]{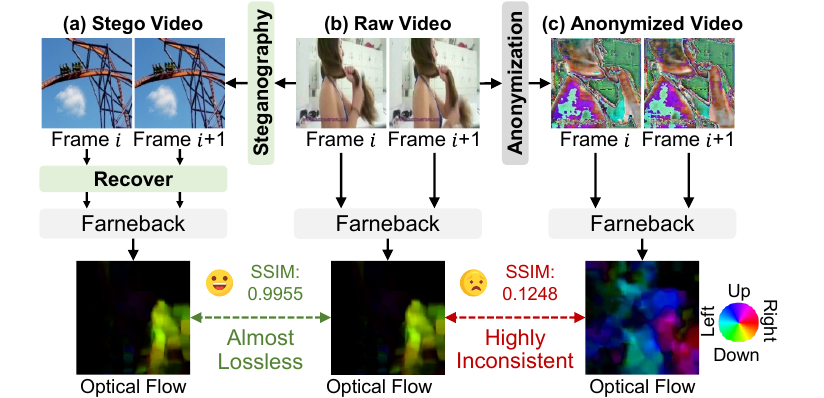}
    \vspace{-1.85em}
	\caption{Comparative optical flow visualization: (a) stego video, (b) raw video, and (c) anonymized video. Raw video is embedded losslessly, maintaining coherent motion patterns, while anonymization disrupts temporal coherence.}
	\label{flow}
\vspace{-1.2em}
\end{figure}

Despite these advancements, prior anonymization techniques share a common flaw: they inevitably alter pixel data, which degrades spatiotemporal integrity and creates visual distortions that risk attracting attackers' attention. Our StegaVAR framework circumvents these issues entirely by losslessly embedding a secret video into a natural-looking cover video and performing analysis directly in the steganographic domain. Throughout both data transmission and action analysis, the hidden secret video remains unexposed. As shown in Fig.~\ref{flow}, anonymization clearly disrupts the temporal information of video, while the process of steganography is lossless. StegaVAR ensures the data remains inconspicuous while preserving original spatiotemporal features, thereby maintaining high performance on the downstream task.

\subsubsection{Media Steganography.}

Traditional steganographic methods offered limited payload capacity by modifying spatial or transform domains \cite{wuVISP2005, panICECC2011, imaizumiIVT2014}. The advent of deep learning revolutionized this field: HiDDeN \cite{Zhu_2018_ECCV} and SteganoGAN \cite{zhang2019steganoganhighcapacityimage} pioneered encoder-decoder structures for image hiding, while Deep Steganography \cite{balujaTPAMI2020} enabled full-size image embedding via convolutional networks. Discrete wavelet transform (DWT) \cite{hePR2012} was subsequently introduced to further enhance the reversibility \cite{dwtstega}. A progress came with Invertible Neural Networks (INNs) \cite{dinh2015nicenonlinearindependentcomponents, dinh2017densityestimationusingreal}, where HiNet \cite{HiNet} established parameter-shared bidirectional mapping for reversible transformations. Building upon image hiding, research has increasingly focused on video hiding, which requires greater embedding capacity \cite{wengICMR2019}. The approach using separate encoder-decoder mechanisms for hiding and recovery operations \cite{Islam_2019} remains effective, but results in high model complexity. LF-VSN \cite{LF_VSN} achieved large-capacity multi-video hiding via a unified INN while reducing model complexity.

\section{Methodology}

\begin{figure*}[t]
	\centering
	\includegraphics[width=\linewidth,scale=1.00]{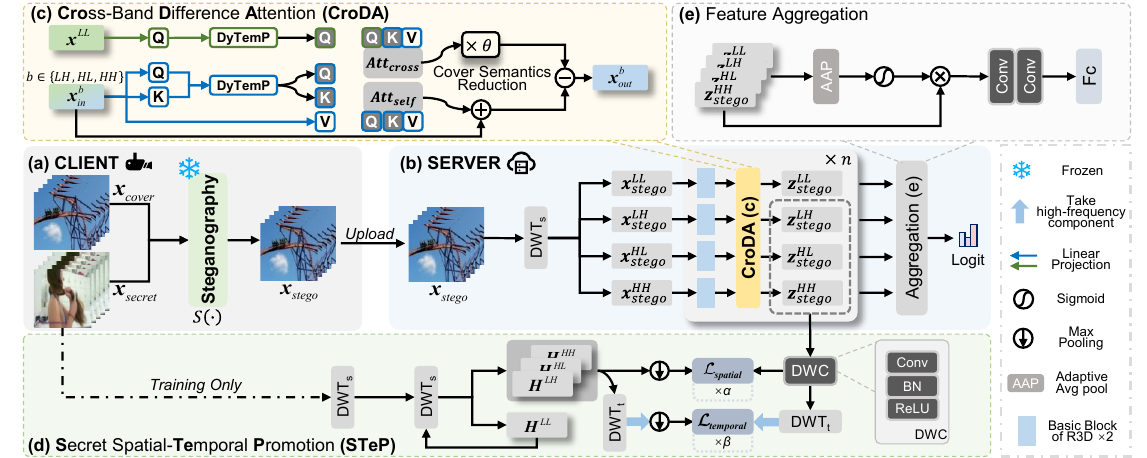}
    \vspace{-2.0em}
	\caption{Illustration of StegaVAR's full pipeline. (a) At the client side, the steganography network $\mathcal{S}$ embeds $\boldsymbol{x}_{secret}$ into $\boldsymbol{x}_{cover}$ to generate $\boldsymbol{x}_{stego}$. (b) On the server side, SDANet decomposes $\boldsymbol{x}_{stego}$ into four sub-bands via DWT and extracts features from each sub-band separately. (c) The CroDA module computes cross-band feature differences to suppress interference from the cover video. (d) During training, the STeP module leverages the original secret video to guide feature learning separately along spatial and temporal dimensions. (e) Final classification is produced through feature aggregation and fully-connected layers.}
	\label{Fig2}
\vspace{-1.2em}
\end{figure*}

Our StegaVAR framework uses a client-side steganography network $\mathcal{S}(\cdot)$ and a server-side steganographic domain analysis network $\mathcal{A}(\cdot)$ for end-to-end privacy-preserving action recognition. As shown in Fig.~\ref{Fig2}, the entire pipeline can be conceptualized as a covert channel communication system. For an action video $\boldsymbol{x}_{secret}$ containing private information, the client embeds it into a visually natural cover video $\boldsymbol{x}_{cover}$ through the steganography network $\mathcal{S}(\cdot)$, generating a stego video containing secret information: $\boldsymbol{x}_{stego} = \mathcal{S}(\boldsymbol{x}_{cover}, \boldsymbol{x}_{secret})$. The stego video $\boldsymbol{x}_{stego}$ is visually indistinguishable from the original cover video $\boldsymbol{x}_{cover}$, avoiding attackers' attention. The server's analysis network, $\mathcal{A}(\cdot)$, then performs steganographic domain analysis on the stego video to yield the prediction $\hat y = \mathcal{A}(\boldsymbol{x}_{stego})$. Throughout this process, the private video $\boldsymbol{x}_{secret}$ always exists in concealed frequency-domain component form and is never exposed in transmission links or server memory. Compared to anonymization-dependent methods, StegaVAR attracts no adversarial attention, eliminating security risks induced by anonymization. Moreover, the server never reconstructs the original video, fundamentally preventing privacy leakage while avoiding disruption of original temporal information.

\subsection{Steganographic Domain Analysis Network}

Since the secret information is primarily embedded in high-frequency components, existing vanilla VAR networks struggle to analyze in the steganographic domain effectively (as demonstrated by experiments in Table.~\ref{VAR_result}). This primarily occurs because the visual representation of $\boldsymbol{x}_{stego}$ mainly originates from $\boldsymbol{x}_{cover}$, while VAR-related features are concealed within the steganographic domain and difficult to extract. Therefore, we decompose $\boldsymbol{x}_{stego}$ using Discrete Wavelet Transform (DWT) to obtain four sub-bands: $\boldsymbol{x}_{stego}^{\textit{LL}}, \boldsymbol{x}_{stego}^{\textit{LH}}, \boldsymbol{x}_{stego}^{\textit{HL}}, \boldsymbol{x}_{stego}^{\textit{HH}} \in \mathbb{R}^{\textit{T} \times \frac{H}{2} \times \frac{W}{2} \times 3}$, representing low-frequency and high-frequency features. Subsequently, we employ ResNet3D-18 \cite{Tran_2018_CVPR} as the backbone to separately analyze these four sub-bands.

To resolve the misalignment between steganographic features and original action semantics, we design the Secret Spatio-Temporal Promotion (STeP) module, which explicitly guides feature learning of each sub-band in $\boldsymbol{x}_{stego}$ during training using spatiotemporal features from $\boldsymbol{x}_{secret}$, enforcing spatial layout and temporal evolution alignment between steganographic representations and secret actions (note that $\boldsymbol{x}_{secret}$ is excluded during inference). Since the cover video's semantics reside mainly in the low-frequency sub-band $\boldsymbol{x}_{stego}^{\textit{LL}}$ \cite{HiNet, hidemia}, 
the Cross-Band Difference Attention (CroDA) module suppresses interference by computing differences between it and the high-frequency sub-bands (LH/HL/HH). CroDA also uses a shared learnable position embedding to ensure temporal consistency across all frequency components. Finally, features from each sub-band are processed by Adaptive Average Pooling, and a two-layer MLP with a sigmoid activation generates weights for each sub-band. Weighted feature aggregation is then performed via a two-layer convolutional network.

\subsubsection{Secret Spatio-Temporal Promotion.}

Steganographic models often embed different parts of the secret video into distinct frequency bands of the cover video using varied methods \cite{hidemia}, which makes directly learning secret features within these bands challenging. Beyond steganography, the Discrete Wavelet Transform (DWT) is also employed to extract high-frequency detail information from images, frequently used in tasks requiring attention to fine-grained texture information such as industrial defect detection \cite{zhangTIP2023}. Inspired by such methods, we introduce an additional STeP branch during training. Unlike traditional applications of DWT in vision tasks, we leverage it not only to decompose spatial features in horizontal and vertical directions but also to further decompose features along the temporal dimension. This decomposition along the temporal dimension allows the module to guide feature learning across temporal scales, which is critical for capturing fine-grained actions defined by subtle motion dynamics. With this spatiotemporal supervision, the network can better capture information in the steganographic domain (as shown in Fig.~\ref{Fig2}).

The STeP branch uses high-frequency spatiotemporal features from $\boldsymbol{x}_{secret}$ to supervise SDANet's learning of high-frequency features in $\boldsymbol{x}_{stego}$. First, we apply a transform to $\boldsymbol{x}_{secret}$ and obtain its $LL$ band to ensure dimensional alignment with subsequent feature maps:
\begin{equation}\small
    H^{LL}_0, H^{LH}_0, H^{HL}_0, H^{HH}_0 = \mathrm{DWT_s}(\boldsymbol{x}_{secret}),
\end{equation}
where $\mathrm{DWT_s(\cdot)}$ denotes the discrete wavelet transform in spatial dimension. Then we perform four levels of DWT on $H^{LL}_0$ (each subsequent transform is applied to the low-frequency component from the previous level). For $n\in\{1,2,3,4\}$:
\begin{equation}\small
H^{LL}_n, H^{LH}_n, H^{HL}_n, H^{HH}_n = \mathrm{DWT_s}(H^{LL}_{n-1}),
\label{eq:multilevel_dwt}
\end{equation}
where $H^{LH}_n, H^{HL}_n, H^{HH}_n \in \mathbb{R}^{T \times \frac{H}{2^{n+1}} \times \frac{W}{2^{n+1}} \times 3}$ denote the high-frequency components at the $n$-th level. Subsequently, apply DWT along the temporal dimension to these components. Since the temporal dimension $T$ is compressed during feature extraction, we use max pooling on the high-frequency components to create the spatial (${G}^s$) and temporal (${G}^t$) ground truth signals for promotion. This process can be represented as:
\begin{equation}\small
\begin{split}
&{G}^{s,b}_n = \mathcal{P}_{n} (H^{b}_n ),\quad b\in \mathcal{B},\\
&{G}^{t,b}_n = \mathcal{P}_{n} (\mathrm{DWT_t}( H^{b}_n )_{high}),
\label{eq:temp_feature}
\end{split}
\end{equation}
where $\mathrm{DWT_t}(\cdot)_{high}$ denotes the wavelet decomposition along the temporal dimension retaining only high-frequency components, and $\mathcal{B}$ denotes the set of three high-frequency sub-bands $\{LH,HL,HH\}$. $\mathcal{P}_{n}$ represents the level-dependent max pooling operation. It only selects significant features along the time dimension without changing height and width. Next, the sub-band features of the stego video are processed by a simple Deep Wavelet Compression (DWC) module (comprising $1\times1$$\times1$ convolution, batch normalization, and ReLU) to align channel dimensions with the high-frequency components of the secret video:
\begin{equation}\small
\begin{split}
    &M^{s,b}_n = \mathrm{DWC}_n \left( z_{n}^{b}  \right),\quad b\in \mathcal{B},\\
    &M^{t,b}_n = \mathrm{DWT_t}(M^{s,b}_n)_{high},
\end{split}
\label{eq:dwc}
\end{equation}
where $z_{n}$ represents the the sub-band feature at the $n$-th layer. Then, calculate the spatial guidance loss $\mathcal{L}_{spatial}$ and the time guidance loss $\mathcal{L}_{temporal}$ respectively:
\begin{equation}\small
    \begin{split}
        &\mathcal{L}_{spatial}^{b} = \sum_{n=1}^N\mathcal{L}_{mse}(G^{s,b}_n,M^{s,b}_n),\\
        &\mathcal{L}_{temporal}^{b} =\sum_{n=1}^N \mathcal{L}_{mse}(G^{t,b}_n,M^{t,b}_n),
    \end{split}
\label{eq:lossst}
\end{equation}
where $N$ represents the four DWT performed and $\mathcal{L}_{mse}$ denotes calculating mean squared error loss. Finally, calculate the total $\mathcal{L}_{STeP}$: 
\begin{equation}\small
\mathcal{L}_{STeP} = \mathcal{\alpha}\cdot \sum_{b\in B} \mathcal{L}_{spatial}^{b} + \mathcal{\beta}\cdot \sum_{b\in B} \mathcal{L}_{temporal}^{b},
\label{eq:lossstep}
\end{equation}
where $\mathcal{\alpha}$ and $\mathcal{\beta}$ respectively represent the strength of spatial and temporal promotion. The whole optimization algorithm with STeP is shown in Algorithm \ref{optimization}.

\subsubsection{Cross-Band Difference Attention.}

We frame the task of VAR in the steganographic domain as a signal denoising problem. The high-frequency sub-band of stego video contains a superposition of the desired action signal from $\boldsymbol{x}_{secret}$ and a small amount of unwanted noise from $\boldsymbol{x}_{cover}$. Considering that the low-frequency sub-band $\boldsymbol{x}_{stego}^{\textit{LL}}$ contains the majority of the cover video's semantic information \cite{HiNet, guanTPAMI2022}, while the secret information is primarily embedded within the three high-frequency sub-bands ($\boldsymbol{x}_{stego}^{\textit{LH}}, \boldsymbol{x}_{stego}^{\textit{HL}}, \boldsymbol{x}_{stego}^{\textit{HH}}$), we compute element-wise difference attention between each high-frequency sub-band and the low-frequency sub-band $\boldsymbol{x}_{stego}^{\textit{LL}}$. This process aims to remove residual cover information lingering within the high-frequency sub-bands, effectively performing a content-adaptive filtering operation that can suppress the cover-related interference and enhance the underlying action signal.

Simultaneously, although the high-frequency sub-bands contain distinct features spatially, their subtle temporal variations are similar. Therefore, we employ two sets of position embeddings to separately encode the low-frequency sub-band and the high-frequency sub-bands.To enhance perception of subtle motion variations, we propose \textbf{Dy}namic \textbf{Tem}poral \textbf{P}erception (\textbf{DyTemP}). In order to adaptively perceive and unify temporal information in different sub-bands, we introduce a learnable position-specific offset based on RoPE \cite{rope}. This hybrid approach preserves relative position awareness while adaptively adjusting the absolute temporal landmarks essential for VAR. Specifically, for an input vector $x \in \mathbb{R}^{T\times d}$ partitioned into $x=\begin{bmatrix}u,v\end{bmatrix}$ where $u,v \in \mathbb{R}^{T\times\frac{d}{2}}$, the DyTemP $E(x)$ is defined as below:
\begin{small}
\begin{gather}
E(x) = \begin{bmatrix} \mathbf{u} \odot (\cos\mathbf{\Theta}+\mathbf{\varepsilon}_{cos}) - \mathbf{v} \odot (\sin\mathbf{\Theta}+\mathbf{\varepsilon}_{sin}) \\ \mathbf{u} \odot (\sin\mathbf{\Theta}+\mathbf{\varepsilon}_{sin}) + \mathbf{v} \odot (\cos\mathbf{\Theta}+\mathbf{\varepsilon}_{cos}) \end{bmatrix},
\end{gather}
\end{small}

\noindent
where $\Theta$ denotes the matrix of rotation angles pre-computed based on token position and feature dimension. $\varepsilon_{\cos}$ and $\varepsilon_{\sin}$ are learnable biases. The implementation of CroDA can be formulated as follows:
\label{eq:rope}
\begin{small}
\begin{gather}
\textrm{CA}(x^{LL},x^{b})=\sigma(\frac{E_{LL}(Q(x^{LL}))E_{b}(K(x^{b}))^\top}{\sqrt{d}})V(x^{b}), \\
\textrm{SA}(x^{b})=\sigma(\frac{E_{b}(Q(x^{b}))E_{b}(K(x^{b}))^\top}{\sqrt{d}})V(x^{b}), \\
x^{b}_{out}=x^{b}_{in}+\textrm{SA}(x^{b}_{in})-\theta \cdot \textrm{CA}(x^{LL},x^{b}_{in}),\quad b \in \mathcal{B},
\label{eq:CroDA}
\end{gather}
\end{small}

\noindent
where $\textrm{CA}(\cdot)$ and $\textrm{SA}(\cdot)$ denote the standard implementation of cross-attention and self-attention. $\sigma$ represents the $\text{Softmax}$ function and $\theta$ is the subtraction strength for the difference component. $E_{LL}$ and $E_{b}$ correspond to position embedding operations applied to the low-frequency and high-frequency sub-bands, respectively, with parameters in $E_{b}$ shared across $LH, HL$ and $HH$. $Q$, $K$, $V$ represent linearly projected values for query, key, and value, respectively.

\begin{algorithm}[t]
\caption{Optimization with STeP}
\label{optimization}
\textbf{Input}: Datasets $\mathcal{D}_{train},\mathcal{D}_{val}$, Pre-trained steganography network $\mathcal{S}$, SDANet $\mathcal{A}$, $N_{epochs}$, hyperparameters $\alpha, \beta, \theta$ \\
\textbf{Output}: Optimal SDANet $\mathcal{A}^*$
\begin{algorithmic}[1]
\STATE Initialize $\mathcal{A}$ with $\theta$
\FOR{epoch $\gets$ 1 to $N_{epochs}$}
    \FOR{each batch $(\boldsymbol{x}_{secret}, \boldsymbol{x}_{cover}, y)$ in $\mathcal{D}_{train}$}
        \STATE $\boldsymbol{x}_{stego} \gets \mathcal{S}(\boldsymbol{x}_{cover}, \boldsymbol{x}_{secret})$
        \STATE $(M^s, M^t, \hat{y}) \gets \mathcal{A}(\boldsymbol{x}_{stego}, \boldsymbol{x}_{secret})$
        \STATE Calculate $\mathcal{L}_{spatial}$ and $\mathcal{L}_{temporal}$ by Eq.~\ref{eq:lossst}
        \STATE Compute classification loss: $\mathcal{L}_{cls} \gets \text{CE}(\hat{y}, y)$
        \STATE Total loss: $\mathcal{L} \gets \mathcal{L}_{cls} + \alpha\cdot\mathcal{L}_{spatial} + \beta\cdot\mathcal{L}_{temporal}$
        \STATE Update $\mathcal{A}$ through back propagation
    \ENDFOR
\ENDFOR
\STATE Obtain the optimal model $\mathcal{A}^*$
\end{algorithmic}
\end{algorithm}

\begin{table*}
\centering
\caption{Comparison with existing privacy-preserving VAR frameworks across datasets, while VAR reports Top-1 (\%) and Privacy reports cMAP (\%) and F1. ↑ denotes higher is better, ↓ denotes lower is better.}
\vspace{-0.8em}
\label{mainresults}
\resizebox{\linewidth}{!}{
  \tabcolsep=18pt
  \renewcommand{\arraystretch}{0.7}
  \begin{tabular}{c|cc|cccc}
  \hline
  \multirow{3}{*}{\textbf{Method}} & \multicolumn{2}{c|}{\rule{0pt}{8pt}\textbf{VAR Performance}}                     & \multicolumn{4}{c}{\textbf{Privacy-Preserving Performance}}                                     \\ \cline{2-7} 
                                   & \multicolumn{1}{c|}{\rule{0pt}{8pt}\textbf{UCF101}}  & \textbf{UCF101→HMDB51} & \multicolumn{2}{c|}{\textbf{VISPR1}}              & \multicolumn{2}{c}{\textbf{VPHMDB}} \\ \cline{2-7} 
                                   & \multicolumn{1}{c|}{\rule{0pt}{8pt}\textbf{Top-1 ↑}} & \textbf{Top-1 ↑}       & \textbf{cMAP ↓} & \multicolumn{1}{c|}{\textbf{F1 ↓}} & \textbf{cMAP ↓}    & \textbf{F1 ↓}    \\ \hline
  \rule{0pt}{8pt}Raw data                         & \multicolumn{1}{c|}{71.98}            & 44.25                  & 64.41           & \multicolumn{1}{c|}{0.555}         & 76.62              & 0.684            \\ \hline
  \rule{0pt}{8pt}Downsample-2×                    & \multicolumn{1}{c|}{54.11}            & 24.10                  & 57.23           & \multicolumn{1}{c|}{0.483}         & 71.35              & 0.601            \\
  Downsample-4×                    & \multicolumn{1}{c|}{39.65}            & 16.80                  & 50.07           & \multicolumn{1}{c|}{\textbf{0.379}}         & 69.79              & 0.594            \\
  Obf-Blackening                   & \multicolumn{1}{c|}{53.13}            & 26.20                  & 56.39           & \multicolumn{1}{c|}{0.457}         & 74.06              & 0.649            \\
  Obf-StrongBlur                   & \multicolumn{1}{c|}{55.59}            & 26.40                  & 55.94           & \multicolumn{1}{c|}{0.456}         & 74.33              & 0.655            \\
  Obf-WeakBlur                     & \multicolumn{1}{c|}{61.52}            & 33.70                  & 63.52           & \multicolumn{1}{c|}{0.523}         & 75.11              & 0.663            \\
  Noise-Features                   & \multicolumn{1}{c|}{61.90}            & 31.20                  & 62.40           & \multicolumn{1}{c|}{0.531}         & -                  & -                \\
  VITA                             & \multicolumn{1}{c|}{62.10}            & 33.20                  & 55.32           & \multicolumn{1}{c|}{0.461}         & 73.89              & 0.638            \\
  SPAct                            & \multicolumn{1}{c|}{62.03}            & 34.10                  & 57.43           & \multicolumn{1}{c|}{0.473}         & -                  & -                \\
  BPAP                             & \multicolumn{1}{c|}{62.11}            & 34.52                  & 57.10           & \multicolumn{1}{c|}{0.450}         & 69.95              & \textbf{0.519}   \\ \hline
  \rule{0pt}{8pt}\textbf{StegaVAR (Weng)}           & \multicolumn{1}{c|}{70.32}            & 42.88                  & 51.66           & \multicolumn{1}{c|}{0.459}         & 59.78              & 0.549            \\
  \textbf{StegaVAR (HiNet)}          & \multicolumn{1}{c|}{70.08}            & 42.75                  & \textbf{47.10}  & \multicolumn{1}{c|}{0.399} & 58.93              & 0.530            \\
  \textbf{StegaVAR (LF-VSN)}         & \multicolumn{1}{c|}{\textbf{71.66}}   & \textbf{43.66}         & 47.87           & \multicolumn{1}{c|}{0.507}         & \textbf{56.65}     & 0.531            \\ \hline
  \end{tabular}%
}
\vspace{-1.0em}
\end{table*}

\section{Experiments}

\subsection{Datasets \& Metrics}
\subsubsection{Datasets.}

For VAR evaluation, we select the two most widely adopted datasets: UCF101 \cite{soomro2012ucf101dataset101human} and HMDB51 \cite{hmdb}. To quantitatively compare privacy preservation performance with other methods, we utilize two subsets of VISPR \cite{Orekondy_2017_ICCV} and the privacy-annotated versions of UCF101 and HMDB51, namely VPUCF101 and VPHMDB51 \cite{STPrivacy} (Further details are provided in Supp.). All datasets employ their official split protocols. We randomly sample 1000 video clips from YouTube-VIS \cite{Yang_2019_ICCV} as cover videos, ensuring no overlap between the training and testing cover videos. For cover samples with fewer than 16 frames, we employ reverse-order padding to extend the sequence length.

\subsubsection{Metrics.}

Following previous works \cite{SPAct, Aslam_2025_CVPR}, we employ Top-1 accuracy to evaluate VAR performance. For privacy preservation performance, we use classwise-mAP (mean Average Precision) and classwise-F1 score.

\subsection{Implementation Details}
\subsubsection{Input Details.}

In all experiments, we first cropped each frame to 0.8× the original dimensions before resizing to an input resolution of 224×224. Each video clip consisted of 16 frames, sampled from random starting points at a frame skip rate of 4. During training, we applied standard augmentations to $\boldsymbol{x}_{secret}$, including random erasing, random cropping, horizontal flipping, and random color jittering.

\subsubsection{Initialization and Training Details.}

The steganography model $\mathcal{S}$ used frozen DIV2K-pretrained weights \cite{agustsson2017ntire}; all VAR models trained from scratch. In StegaVAR, $\theta$ in Eq.~\ref{eq:CroDA} was set to 0.2. For raw data or other privacy methods, $\theta=0$ retained only self-attention. This hyperparameter setting is based on CroDA's function of removing low-frequency information from high-frequency components, a difference computation strategy that is uniquely effective in the steganographic domain and would otherwise degrade performance. VAR models trained 150 epochs with Adam (lr=1e-4, batch=32), loss coefficients $\alpha$=0.2 and $\beta$=0.3 in Eq.~\ref{eq:lossstep}. Privacy evaluation used ResNet-50 \cite{He_2016_CVPR} with ImageNet weights \cite{dengCVPR2009} trained 100 epochs under identical optimization. All experiments were implemented in PyTorch on four NVIDIA RTX 4090 GPUs.

\begin{table}[!t]
\centering
\caption{Comparison of different privacy preserving methods and VAR network (our SDANet/ResNet3D) combinations, evaluated using Top-1 (\%).}
\label{VAR_result}
\vspace{-0.8em}
\resizebox{\linewidth}{!}{
\renewcommand{\arraystretch}{0.7}
\begin{tabular}{cc|c|c}
\hline
\multicolumn{2}{c|}{\textbf{Method}}                                    & \textbf{UCF101}  & \textbf{\begin{tabular}[c]{@{}c@{}}\rule{0pt}{8pt}UCF101→\\ HMDB51\end{tabular}} \\ \hline
\multicolumn{1}{c|}{\textbf{\rule{0pt}{8pt}Privacy-preserving}} & \textbf{VAR Network} & \textbf{Top-1 ↑} & \textbf{Top-1 ↑}                                                  \\ \hline
\multicolumn{1}{c|}{\multirow{2}{*}{Raw data}}   & \rule{0pt}{8pt}ResNet3D             & 62.33            & 35.60                                                             \\
\multicolumn{1}{c|}{}                            & SDANet               & 71.98            & 44.25                                                             \\ \hline
\multicolumn{1}{c|}{\multirow{2}{*}{BPAP}}       & \rule{0pt}{8pt}ResNet3D             & 62.11            & 34.52                                                             \\
\multicolumn{1}{c|}{}                            & SDANet               & 61.22            & 37.81                                                             \\ \hline
\multicolumn{1}{c|}{\rule{0pt}{8pt}Weng}                        & ResNet3D             & 59.08            & 33.13                                                             \\
\multicolumn{1}{c|}{HiNet}                       & ResNet3D             & 58.69            & 33.28                                                             \\
\multicolumn{1}{c|}{LF-VSN}                      & ResNet3D             & 58.88            & 32.67                                                             \\ \hline
\multicolumn{1}{c|}{\rule{0pt}{8pt}\textbf{StegaVAR (Weng)}}      & SDANet               & 70.32            & 42.88                                                             \\
\multicolumn{1}{c|}{\textbf{StegaVAR (HiNet)}}     & SDANet               & 70.08   & 42.75                                                    \\
\multicolumn{1}{c|}{\textbf{StegaVAR (LF-VSN)}}    & SDANet               & \textbf{71.66}   & \textbf{43.66}                                                    \\ \hline
\end{tabular}
}
\vspace{-1.2em}
\end{table}

\subsection{Main Results}

\subsubsection{VAR Performance.}

As shown in Table.~\ref{mainresults}, our StegaVAR framework significantly outperforms existing privacy-preserving methods, achieving accuracy comparable to non-private baselines. Specifically, StegaVAR with LF-VSN \cite{LF_VSN} achieves 71.66\% Top-1 accuracy on UCF101 and 43.66\% on HMDB51, surpassing the state-of-the-art BPAP \cite{Aslam_2025_CVPR} by over 9\% on both datasets while incurring only a minimal 0.32\%/0.59\% drop compared to the raw data baseline. The superiority of LF-VSN over other steganographic models like Weng \cite{wengICMR2019} and HiNet \cite{HiNet} is attributed to its use of inter-frame information, which better preserves temporal features. Furthermore, the poor performance of vanilla VAR networks on stego videos (shown in Table.~\ref{VAR_result}) confirms the necessity of a specialized network for effective analysis within the steganographic domain.

Notably, in Table.~\ref{VAR_result}, SDANet substantially outperforms the vanilla ResNet3D on raw data (71.98\% v.s. 62.33\% on UCF101). This advantage stems from the wavelet decomposition in STeP, where high-frequency components provide fine-grained supervisory signals that enhance focus on subtle temporal variations. Its effectiveness outside the steganographic domain confirms the cross-domain universality of this guidance mechanism. Conversely, SDANet's performance drops to 61.22\% on UCF101 when applied to the anonymizing BPAP framework, falling below ResNet3D. This failure demonstrates the inability of DWT to extract meaningful signals, which validates that anonymization techniques inherently cause irreversible spatiotemporal disruption, whereas our steganography-based approach preserves video integrity through embedding.

\begin{table}[!t]
\centering
\caption{Comparison of privacy-preserving performance across datasets, evaluated using cMAP (\%) and F1 score.}
\vspace{-0.8em}
\label{privacy_result}
\resizebox{\linewidth}{!}{
\renewcommand{\arraystretch}{0.7}
\begin{tabular}{c|cc|cc}
\hline
\multirow{2}{*}{\textbf{Method}} & \multicolumn{2}{c|}{\textbf{\rule{0pt}{8pt}VISPR1→2}} & \multicolumn{2}{c}{\textbf{VPHMDB→VPUCF}} \\ \cline{2-5} 
                                 & \textbf{\rule{0pt}{8pt}cMAP ↓}    & \textbf{F1 ↓}     & \textbf{cMAP ↓}      & \textbf{F1 ↓}      \\ \hline
\rule{0pt}{8pt}Raw data                         & 57.63              & 0.434             & 76.62                & 0.699              \\ \hline
\rule{0pt}{8pt}VITA                             & 49.60              & 0.399             & 76.02                & 0.669              \\
SPAct                            & 47.10              & 0.386             & 75.98                & 0.661              \\
BPAP                             & 49.50              & 0.352             & 70.91                & 0.612              \\ \hline
\textbf{\rule{0pt}{8pt}StegaVAR (Weng)}           & \textbf{45.93}     & \textbf{0.347}    & 62.54                & 0.532              \\
\textbf{StegaVAR (HiNet)}          & 46.84              & 0.391             & \textbf{61.74}       & 0.531              \\
\textbf{StegaVAR (LF-VSN)}         & 47.04              & 0.389             & 61.83                & \textbf{0.526}     \\ \hline
\end{tabular}
}
\vspace{-1.2em}
\end{table}

\subsubsection{Privacy-preserving.}

Although our primary objective is covert transmission and steganographic domain analysis, we quantitatively evaluate StegaVAR's privacy protection by extracting features from $\boldsymbol{x}_{stego}$ using ResNet-50 \cite{He_2016_CVPR}. As shown in Table.~\ref{mainresults}, ResNet-50 achieves significantly lower privacy recognition on StegaVAR-processed videos, attaining 47.10\% cMAP and 0.399 F1 on VISPR1. This outperforms the strongest competitor (55.32\% cMAP / 0.461 F1) by 8.22\% (cMAP$\downarrow$) and 0.062 (F1$\downarrow$). StegaVAR also leads in transfer tasks: StegaVAR (Weng) achieves 45.93\% cMAP/0.347 F1 on VISPR1$\rightarrow$VISPR2, while StegaVAR (HiNet) attains 61.74\% cMAP on VPHMDB$\rightarrow$VPUCF (Table.~\ref{privacy_result}). This demonstrates effective resistance against automated privacy inference attacks beyond human perception deception.

\begin{table*}[t]
\centering
\begin{minipage}[t]{0.38\textwidth}
    \centering
    \caption{Ablation results of STeP and CroDA on UCF101, fixing $\alpha$=0.2, $\beta$=0.3, $\theta$=0.2.}
    \label{module}
    \vspace{-0.8em}
    \resizebox{\linewidth}{!}{
        \renewcommand{\arraystretch}{0.33}
        \begin{tabular}{ccc|c}
            \hline
            \textbf{\begin{tabular}[c]{@{}c@{}}\rule{0pt}{9pt}Spatial\\ Promotion\end{tabular}} & \textbf{\begin{tabular}[c]{@{}c@{}}Temporal\\ Promotion\end{tabular}} & \textbf{CroDA} & \textbf{Top-1 (\%) ↑} \\ \hline
            \rule{0pt}{8pt}\textbf{×}                                                              & \textbf{×}                                                               & \textbf{×}     & 63.15                  \\
            $\checkmark$                                                                          & \textbf{×}                                                               & \textbf{×}     & 66.29                  \\
            \textbf{×}                                                              & $\checkmark$                                                               & \textbf{×}     & 66.16                  \\
            \textbf{×}                                                              & \textbf{×}                                                               & $\checkmark$   & 65.81                  \\
            $\checkmark$                                                                          & $\checkmark$                                                               & \textbf{×}     & 68.54                  \\
            $\checkmark$                                                                          & \textbf{×}                                                               & $\checkmark$   & 68.86                  \\
            \textbf{×}                                                              & $\checkmark$                                                               & $\checkmark$   & 68.31                  \\ \hline
            \rowcolor{gray!15}
            \rule{0pt}{8pt}$\checkmark$                                                                          & $\checkmark$                                                               & $\checkmark$   & \textbf{71.66}         \\ \hline
        \end{tabular}
    }
\end{minipage}
\hfill 
\begin{minipage}[t]{0.28\textwidth}
    \centering
    \caption{Ablation results of position embedding strategies on UCF101.}
    \label{PE}
    \vspace{-0.8em}\resizebox{\linewidth}{!}{
        \renewcommand{\arraystretch}{1.55}
    \begin{tabular}{c|c}
        \hline
        \textbf{Position Embedding} & \textbf{Top-1 (\%) ↑} \\ \hline
        w/o PE                      & 70.39                  \\
        Absolute PE                 & 70.64                  \\
        RoPE                        & 71.03                  \\ \hline
        \rowcolor{gray!15}
        DyTemP                      & \textbf{71.66}         \\ \hline
    \end{tabular}
    }
\end{minipage}
\hfill 
\begin{minipage}[t]{0.32\textwidth}
    \centering
    \caption{Ablation results of sub-band grouping strategies on UCF101.}
    \label{group}
    \vspace{-0.8em}\resizebox{\linewidth}{!}{
        \renewcommand{\arraystretch}{1.09}
    \begin{tabular}{l|c}
        \hline
        \textbf{Group of sub-bands}      & \textbf{Top-1 (\%) ↑} \\ \hline
        \{\textit{LL, LH, HL, HH}\}        & 58.03                  \\
        \{\textit{LL}\}, \{\textit{LH, HL, HH}\} & 62.73                  \\
        \{\textit{LL}\}, \{\textit{LH}\}, \{\textit{HL, HH}\} & 66.68                  \\
        \{\textit{LL}\}, \{\textit{LH, HL}\}, \{\textit{HH}\} & 66.74                  \\
        \{\textit{LL}\}, \{\textit{HL}\}, \{\textit{LH, HH}\} & 66.27                  \\ \hline
        \rowcolor{gray!15}
        \{\textit{LL}\}, \{\textit{LH}\}, \{\textit{HL}\}, \{\textit{HH}\} & \textbf{71.66}         \\ \hline
    \end{tabular}
    }
\end{minipage}
\end{table*}

\begin{figure}[t]   
\vspace{-1.2em}
	\centering
	\includegraphics[width=\linewidth,scale=1.00]{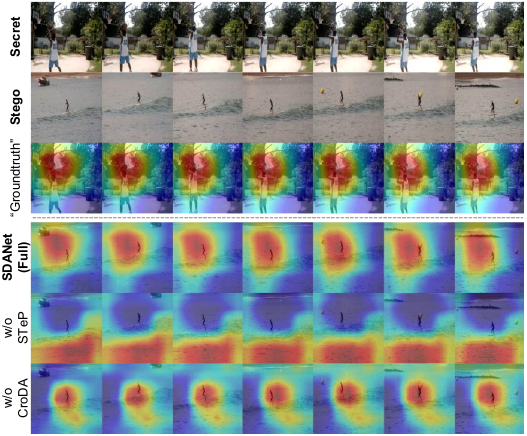}
    \vspace{-1.8em}
	\caption{Visual attention of module ablation.}
	\label{ablation_att}
\vspace{-1.5em}
\end{figure}

\subsection{Ablation Studies}

\subsubsection{Effectiveness of STeP.}

Starting from a baseline accuracy of 63.15\% on UCF101 without STeP or CroDA (Table.~\ref{module}), the spatial promotion component alone improves performance to 66.29\%, while the temporal promotion component achieves 66.16\%. Integrating both spatial and temporal promotion yields 68.54\% accuracy. When these components are individually combined with CroDA, performance is further improved to 68.86\% (spatial promotion + CroDA) and 68.31\% (temporal promotion + CroDA), respectively. These results confirm that explicit guidance from the secret video's high-frequency components is crucial. Furthermore, attention map visualizations in Fig.~\ref{ablation_att} reveal STeP's critical role in action region localization, which is absent without it.

\subsubsection{Effectiveness of CroDA.}

CroDA contributes substantially to suppressing cover-induced interference. As shown in Table.~\ref{module}, standalone CroDA improves baseline performance by 2.66\% (65.81\% vs. 63.15\%). Further analysis reveals that position embedding (PE) is essential for modeling temporal relationships. Models without PE achieve only 70.39\% accuracy (Table.~\ref{PE}). Absolute PE \cite{devlin-etal-2019-bert} marginally improves performance to 70.64\% (+0.25\%), while RoPE \cite{rope} significantly enhances accuracy to 71.03\% by capturing relative temporal dependencies. DyTemP further increases accuracy to 71.66\% due to the learnable offset $\varepsilon$. The visual evidence in Fig.~\ref{ablation_att} underscores CroDA's essential function in suppressing cover interference. Ultimately, integrating all components achieves the highest accuracy, confirming that optimal performance relies on the synergy between STeP's spatio-temporal guidance and CroDA's cross-band attention.

\subsubsection{Group of sub-bands.}
To validate the necessity of analysis for the four distinct frequency sub-bands using four separate ResNet3D, we conducted experiments with different sub-band grouping strategies (Table.~\ref{group}). Concatenating all sub-bands for single-branch processing (retaining only CroDA's self-attention) yields the lowest VAR performance (58.03\%). Significant improvement occurs when isolating the LL sub-band from others (62.73\% v.s. 58.03\%). While three-group configurations achieve comparable results across approaches, independent processing of all four sub-bands peaks at 71.66\% accuracy. This demonstrates substantial spectral divergence of secret information across frequency bands and confirms that separate analysis optimally captures band-specific discriminative features.

\begin{figure}[t]   
\vspace{-1.2em}
	\centering
	\includegraphics[width=\linewidth,scale=1.00]{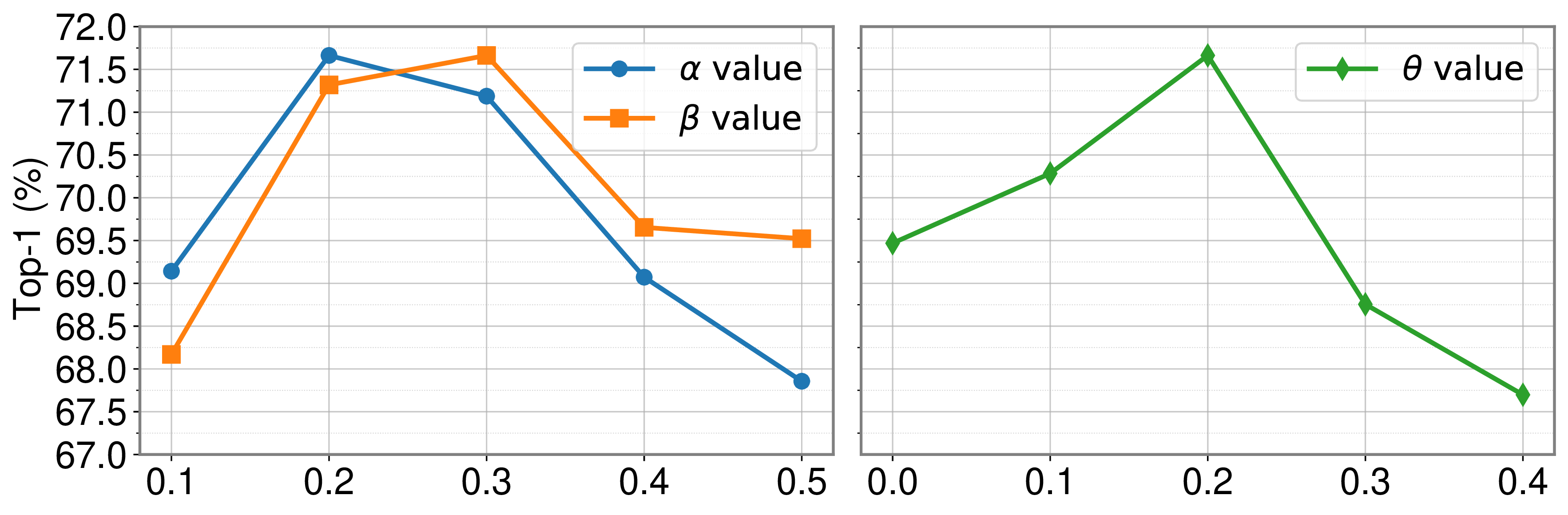}
    \vspace{-1.8em}
	\caption{Effect of the three hyperparameters on UCF101. $\mathcal{\alpha}$ and $\mathcal{\beta}$ respectively represent the strength of spatial and temporal guidance. $\theta$ is the subtraction strength for the difference component of CroDA.}
	\label{hyper}
\vspace{-1.5em}
\end{figure}

\subsection{Hyperparameter Analysis}

Model sensitivity to $\alpha$, $\beta$ (in Eq.~\ref{eq:lossstep}), and $\theta$ (in Eq.~\ref{eq:CroDA}) necessitates strict calibration at optimal values (0.2, 0.3, 0.2). Fixing $\beta$=0.3 and $\theta$=0.2, $\alpha$=0.2 yields peak accuracy (71.66\%), while $\alpha$=0.1 (69.14\%) or $\alpha$=0.5 (67.86\%) cause significant degradation. Similarly, with $\alpha$=0.2 and $\theta$=0.2, $\beta$=0.3 achieves maximum performance (71.66\%), though $\beta$=0.2 remains competitive (71.32\%); values beyond $\beta$=0.3 or below 0.2 reduce accuracy by 2-3.5\%. Notably, $\theta$ exhibits an extremely narrow optimum: $\theta$=0 degrades performance to 69.46\% versus $\theta$=0.2 (71.66\%), proving cross-band difference computation's essential role. Minor deviations to $\theta$=0.1 (70.28\%) or 0.3 (68.76\%) substantially harm accuracy, indicating precise calibration is required for limited low-frequency interference in high-frequency components.

\section{Conclusion}

We propose StegaVAR, a pioneering framework that rethinks privacy-preserving VAR by integrating steganographic principles. By embedding secret videos into cover videos and performing analysis directly in the steganographic domain, our strategy ensures content concealment while avoiding the spatiotemporal disruption of other methods. Experimental results testified that our STeP and CroDA modules effectively guide feature learning and suppress interference, enabling StegaVAR to achieve recognition performance rivaling non-private baselines with robust privacy protection against both human and automated attacks.

While StegaVAR demonstrates significant advantages over anonymization methods, it incurs minor information loss compared to raw video analysis. Future work should explore advanced reversible transformations or adaptive fusion mechanisms to bridge this fidelity gap.

\section{Acknowledgements}
This work was supported by Guangdong Research Team for Communication and Sensing Integrated with Intelligent Computing (Project No. 2024KCXTD047). The computational resources are supported by SongShan Lake HPC Center (SSL-HPC) in Great Bay University.

\bibliography{aaai2026}

\clearpage
\twocolumn[
    \centering
    \Large \textbf{Supplementary Material: StegaVAR: Privacy-Preserving Video Action Recognition via Steganographic Domain Analysis} \\
    \vspace{0.5em}
    \vspace{1.0em}
]

\section{Datasets}
\textbf{UCF101} \cite{soomro2012ucf101dataset101human} is a widely-used large-scale benchmark for video action recognition, comprising 13,320 videos across 101 distinct daily human activities. All experiments in this paper utilize official split 1, containing 9,537 training videos and 3,783 test videos.

\noindent
\textbf{HMDB51} \cite{hmdb} is comparatively smaller, with 6,849 videos spanning 51 human actions. Results are reported using official split 1, consisting of 3,570 training videos and 1,530 test videos.

\noindent
\textbf{VISPR} \cite{Orekondy_2017_ICCV} is a multi-category classification dataset designed for privacy attribute recognition, containing 22,167 images annotated with 68 privacy attributes across dimensions including facial features, gender, skin tone, ethnicity, and nudity. We employ two independent subsets in experiments: VISPR1 and VISPR2, each containing seven distinct privacy attributes (detailed in Table~\ref{pri_label}). Each privacy attribute is treated as a binary label, where 0 indicates absence and 1 indicates presence of the attribute. Classification for individual images is formulated as a multi-label classification task, where multiple privacy attributes may coexist within a single image.

\noindent
\textbf{VPUCF and VPHMDB} \cite{STPrivacy} are large-scale annotated datasets for action recognition tasks, featuring comprehensive privacy attribute annotations. The VPUCF dataset extends the UCF101 benchmark, encompassing 101 human action categories across 13,320 videos. Similarly, VPHMDB derives from HMDB51, containing 51 action categories and 6,849 videos. All videos are annotated with five privacy attributes using binary labeling: face, skin color, gender, nudity, and familial relationship, where 1 indicates attribute presence and 0 denotes absence.

\begin{table}[b]
\centering
\caption{Privacy attributes of VISPR subsets.}
\label{pri_label}
\begin{tabular}{cc}
\hline
\textbf{VISPR1}    & \textbf{VISPR2}    \\ \hline
a17\_color         & a6\_hair\_color    \\
a4\_gender         & a16\_race          \\
a9\_face\_complete & a59\_sports        \\
a10\_face\_partial & a1\_age\_approx    \\
a12\_semi\_nudity  & a2\_weight\_approx \\
a64\_rel\_personal & a73\_landmark      \\
a65\_rel\_soci     & a11\_tattoo        \\ \hline
\end{tabular}
\end{table}

\begin{figure}[t]   
\vspace{-1.2em}
	\centering
	\includegraphics[width=\linewidth,scale=1.00]{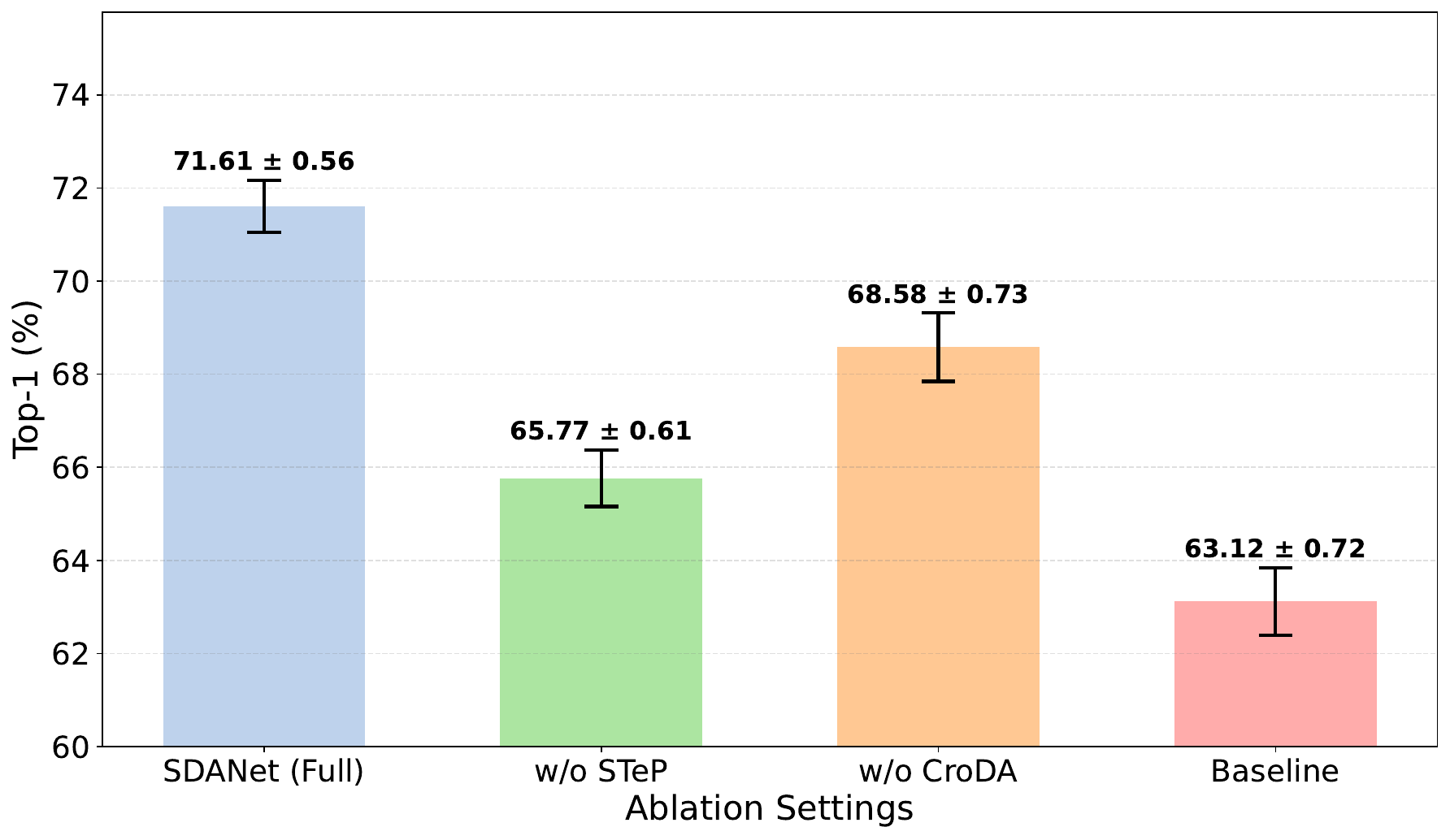}
    \vspace{-1.8em}
	\caption{Sensitivity to cover on UCF101. }
	\label{cover}
\vspace{-1.5em}
\end{figure}

\section{Additional Results}

\subsection{Sensitivity to Cover}

To assess the model's sensitivity to the cover video, an experiment was run using a single, fixed cover for the entire UCF101 test set. As shown in Fig~\ref{cover}, ten repeated trials yielded comparable Top-1 accuracy to variable-cover experiments. The removal of the CroDA module resulted in a marked increase in performance volatility, with the standard deviation increasing significantly ($\pm0.73\%$), demonstrating CroDA's critical role in suppressing cover-induced interference. The complete SDANet achieved minimal standard deviation ($\pm0.56\%$), confirming that synergistic module integration enhances robustness against cover variations.

\subsection{Computational Cost}

Considering our client-server framework where client devices typically have limited computational resources, we compare the FLOPs and parameters of common anonymizers \cite{SPAct, Aslam_2025_CVPR} versus steganographic models. As shown in Table \ref{flops}, Weng's method \cite{wengICMR2019} requires only 36.23G FLOPs, which is significantly lower than UNet's 61.55G \cite{unet} and nearly an order of magnitude below UNet++'s 306.55G \cite{unet++}. Although Weng has higher parameter counts, its low FLOPs indicate superior suitability for computation-constrained client deployment. HiNet \cite{HiNet} achieves the lowest parameter count (4.05M). While its FLOPs exceed Weng's, HiNet remains competitive compared to other methods. LF-VSN exhibits higher FLOPs and parameters due to its utilization of inter-frame auxiliary information during embedding, but this design preserves richer temporal features of secret videos in the steganographic domain, enabling its highest VAR performance. Crucially, our framework's compatibility with diverse steganographic models allows hardware-appropriate selection without imposing greater computational burdens than anonymization approaches, and may even outperform them in resource-constrained scenarios.

\begin{table}[ht]
\centering
\caption{Comparison of FLOPs (G), parameters (M) and Top-1 (\%)(on UCF101) for different privacy-preserving methods.}
\vspace{-0.8em}
\label{flops}
\begin{tabular}{c|cc|c}
\hline
\textbf{Network} & \textbf{FLOPs ↓} & \textbf{Params. ↓} & \textbf{Top-1 ↑} \\ \hline
UNet             & 61.55            & 17.27              & 62.03            \\
UNet++           & 306.55           & 47.20              & 62.11            \\ \hline
Weng             & \textbf{36.23}   & 41.83              & 70.32            \\
HiNet            & 101.44           & \textbf{4.05}      & 70.08            \\
LF-VSN           & 180.51           & 7.20               & \textbf{71.66}   \\ \hline
\end{tabular}
\vspace{-1.0em}
\end{table}

\section{Minor Limitations}
In the main manuscript, we discuss the necessity of reducing information loss during concealment. Here we further identify two minor issues for real-world deployment.

\subsubsection{Extension to Other Video Tasks.}

The current PRIVAS framework is designed for tasks within a client-server architecture, where the server performs its analysis and directly outputs a final result, such as a classification label. This paradigm is not directly applicable to client-server-client tasks, where processed video data might need to be transmitted back to a client. During this return transmission, the data could potentially be exposed. Future work could involve enhancing the framework to accommodate the requirements of client-server-client workflows, ensuring data remains secure throughout the entire process.

\subsubsection{Resistance to Steganalysis.}

Existing steganographic networks may be susceptible to attacks from dedicated steganalysis networks. While the key-controlling design of the LF-VSN \cite{LF_VSN} provides a degree of defense against such attacks, it is necessary to further enhance the security of the steganographic process in the future. Integrating more sophisticated encryption techniques into the steganography pipeline would be a crucial step to bolster the framework's overall security and resilience.

\begin{figure}[t]   
	\centering
	\includegraphics[width=\linewidth,scale=1.00]{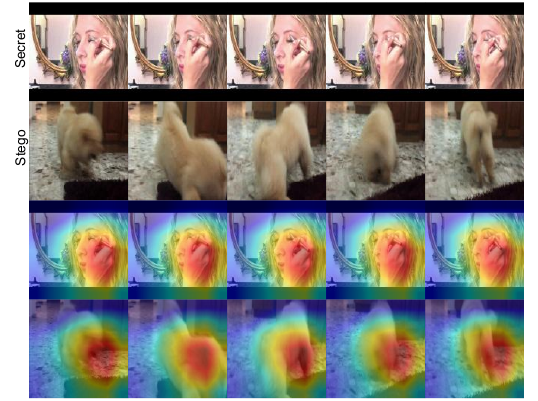}
	\caption{Visual attention of a static secret video and a dynamic cover video.}
	\label{cover_att1}
\end{figure}

\begin{figure}[t]   
	\centering
	\includegraphics[width=\linewidth,scale=1.00]{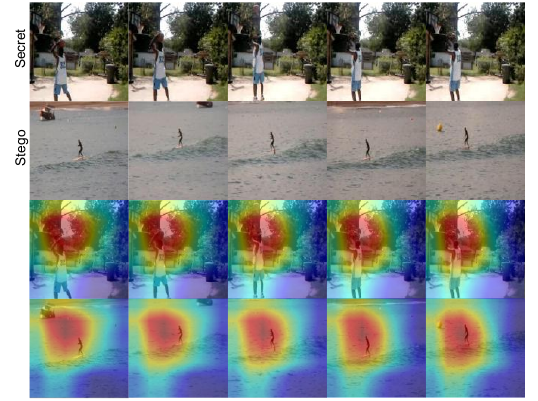}
	\caption{Visual attention of a dynamic secret video and a dynamic cover video.}
	\label{cover_att2}
\end{figure}

\section{Visualization}

\subsection{Influence of Cover}

As illustrated in Fig~\ref{cover_att1}, the secret video depicting "apply eye makeup" exhibits subtle motions while the cover video displays strong dynamic activity, yet the model's visual attention consistently focuses on the hand performing the action without cover interference. Conversely, Fig~\ref{cover_att2} shows the highly dynamic "basketball" action in the secret video alongside vigorous motion in the cover content, but the attention maps correctly localize the executing arm region while rejecting cover distractions. Crucially, the model maintains robust focus across these divergent scenarios because training incorporates diverse cover videos spanning static to dynamic content, where STeP provides precise spatial-temporal guidance to direct attention toward authentic action regions while CroDA actively suppresses cover-induced interference, collectively minimizing performance degradation from cover content.

\begin{figure*}[t]   
	\centering
	\includegraphics[width=\linewidth,scale=1.00]{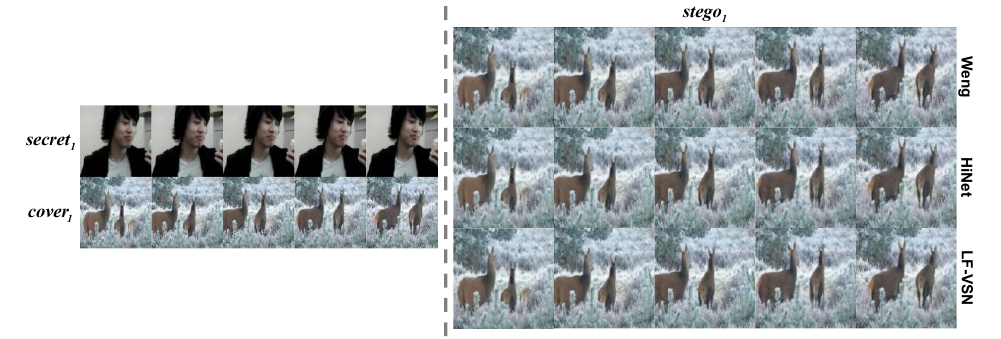}
    \vspace{-1.8em}
	\caption{Visual comparison for different steganographic models.}
	\label{com1}
\end{figure*}

\begin{figure*}[t]   
	\centering
	\includegraphics[width=\linewidth,scale=1.00]{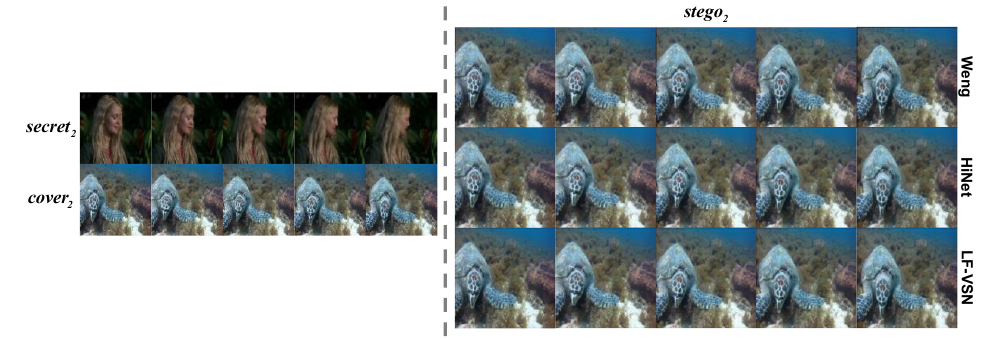}
    \vspace{-1.8em}
	\caption{Visual comparison for different steganographic models.}
	\label{com2}
\end{figure*}

\end{document}